\documentclass{article}

\usepackage{arxiv}

\usepackage[utf8]{inputenc} 
\usepackage[T1]{fontenc}    
\usepackage{hyperref}       
\usepackage{url}            
\usepackage{booktabs}       
\usepackage{amsfonts}       
\usepackage{nicefrac}       
\usepackage{microtype}      
\usepackage{lipsum}
\usepackage{CJKutf8}
\usepackage{textcomp}

\title{LSICC: A Large Scale Informal Chinese Corpus}

\author{
      Jianyu Zhao\\
  School of Data Science\\
  Fudan University, China\\
  \texttt{18210980101@fudan.edu.cn} \\
  \texttt{zhaojianyu@gsqtec.com} \\
   \And
 Zhuoran Ji \\
  GSQ Tec.\\
  Shen Zhen, China \\
  \texttt{jizhuoran@gsqtec.com} \\
}

\begin{document}
\maketitle

\begin{abstract}
Deep learning based natural language processing model is proven powerful, but need large-scale dataset. Due to the significant gap between the real-world tasks and existing Chinese corpus, in this paper, we introduce a large-scale corpus of informal Chinese. This corpus contains around 37 million book reviews and 50 thousand netizen's comments to the news. We explore the informal words frequencies of the corpus and show the difference between our corpus and the existing ones. The corpus can be further used to train deep learning based natural language processing tasks such as Chinese word segmentation, sentiment analysis.
\end{abstract}


\section{Introduction}

Deep learning has been the mainstay for natural language processing, ranging from text summarization \cite{paulus2017deep} to sentiment analysis \cite{zhang2018deep} to text generation \cite{sutskever2014sequence} and automated question-answering system \cite{yu2014deep}. Unlike traditional rule-based methods, the scale and quality of the corpus significantly influence the performance of the deep learning models. In Chinese NLP field, there are many famous large-scale corpora with high quality, such as Baidu Encyclopedia, People's Daily News and Sina Weibo News. Various powerful Chinese deep learning models are trained on these corpora \cite{li2018analogical}, \cite{min2015bosonnlp}, \cite{cui2016consensus}, \cite{nallapati2016abstractive}, \cite{gu2016incorporating}. 

However, most Chinese corpora are in written Chinese, while most real-world deep learning based NLP systems deal with informal Chinese, such as products reviews, netizens' opinions, and microblogs. There are great gaps between informal Chinese and written Chinese, especially in words usages and sentences structures. The pre-trained deep learning model trained from written Chinese corpus, such as words embedding and Chinese words segmentation tools, may perform badly on tasks with informal Chinese.

To address this issue, we introduce LSICC, a large-scale corpus of informal Chinese. Containing around 37 million book reviews and 50 thousand netizens' opinions to news, LSICC is a typical informal Chinese corpus. Most sentences of LSICC are in spoken Chinese and even Internet slang. As far as we know, LSICC is the first large-scale, well-formatted, cleansed corpus focusing on informal Chinese. 

This paper makes the following contributions:
\begin{enumerate}
    \item collect a large scale corpus of informal Chinese
    \item filter out the informationless data items
    \item compare the proportions of informal words in several corpus
\end{enumerate}

\section{Informal Chinese}

Informal Chinese, including spoken Chinese and Chinese Internet Slang, has a substantial difference with the formal one, in both grammar and words usage. In this section, we discuss the difference between formal Chinese and informal Chinese.

\subsection{Spoken Chinese}

For most language, there are differences between the spoken one and the written one. In Chinese, the gap is even more significant due to the long history of written Chinese. 

Similar to another language, spoken Chinese sometimes does not follow the rules as strictly as written Chinese, especially for the elliptical sentences. For example, in spoken Chinese, the subjects sometimes are omitted.

In addition to the grammar, the usage of the words influences the neural network based Chinese natural language processing model most. There are various interchangeable words pairs between written Chinese and spoken Chinese, such as \begin{CJK*}{UTF8}{gbsn}“脑袋”\end{CJK*} and \begin{CJK*}{UTF8}{gbsn}"头部"\end{CJK*}, which both mean ``head'' in Chinese. The two words in each interchangeable words pair usually have almost the same meanings, but the one in written Chinese is more formal, while the one in spoken Chinese is informal.

\subsection{Internet Slang}

Born in the 1990s, Chinese Internet slang refers to various kinds of slang created by netizens and used in chat rooms, social networking services, and online community. Nowadays, Chinese Internet slang is not little memes within internet ingroup, but becoming popular language style of all Chinese speakers. From 2012, Xinhuanet selects ``Top 10 Chinese Internet Slang'' \cite{topten} every year, and Chinese Internet slang is used even by Chinese official institutions.

The first kind of Internet slang is the phonetic substitution, whose pronunciation is same or similar to the formal phrase. For example, in Internet slang, people may use \begin{CJK*}{UTF8}{gbsn}“神马”\end{CJK*} to replace \begin{CJK*}{UTF8}{gbsn}“什么”\end{CJK*}. Both \begin{CJK*}{UTF8}{gbsn}“神马”\end{CJK*} and \begin{CJK*}{UTF8}{gbsn}“什么”\end{CJK*} are pronounced as "cien ma" and has the meaning of ``what''. However, in written Chinese, \begin{CJK*}{UTF8}{gbsn}“神马”\end{CJK*} means ``horse-god'', while \begin{CJK*}{UTF8}{gbsn}“什么”\end{CJK*} means "what".

Transliteration is also a primary way to form Internet slang. As the words are transliterated from another language, both the meaning and pronunciation of the transliterated words are similar to the source language. For example, \begin{CJK*}{UTF8}{gbsn}“伐木累”\end{CJK*} is transliterated from English word ``family'' and only used as Chinese Internet slang \cite{Li:2008:MMR:1613715.1613849}.

Meanwhile, Internet slang is also created by giving new meanings to the old words. For example, in written Chinese, \begin{CJK*}{UTF8}{gbsn}“酱油”\end{CJK*} means ``soy s sauce''. However, in the Chinese Internet slang, it refers to ``passing by". 

\section{Data Collection}

LSICC collects book reviews from DouBan Dushu and netizen's opinions from Chiphell. This section describes these two datasets and pre-processing methods briefly.

\subsection{DouBan DuShu}
DouBan DuShu\footnote{available on: https://github.com/JaniceZhao/Douban-Dushu-Dataset.git} is a Chinese website where users can share their reviews about various kinds of books. Most of the users on this website are unprofessional book reviewers. Therefore, the comments are usually spoken Chinese or even Internet slang. In addition to the comments, users can mark the books from one star to 5 stars according to the quality of the books. We have collected more than 37 million short comments from about 18 thousand books with 1 million users. The great number of users provide diversities of the language styles, from moderate formal to informal. An example of the data item is shown in table \ref{douban}.

\begin{table}[h!]
\centering
\begin{tabular}{l|l|l}
\hline
Key       & Description                                      & Value Example \\ \hline
Book Name & The name of the book                             & \begin{CJK*}{UTF8}{gbsn}理想国\end{CJK*}           \\
User Name & Who gives the comment (anonymized)               & 399           \\
Tag       & The tag the book belongs to                      & \begin{CJK*}{UTF8}{gbsn}思想\end{CJK*}            \\
Comment   & Content of the comment                           & \begin{CJK*}{UTF8}{gbsn}我是国师的脑残粉\end{CJK*}      \\
Star      & Stars given to the book (from 1 star to 5 stars) & 5 stars       \\
Date      & When the comment posted                          & 2018-08-21    \\
Like      & Count of ``like'' on the comment                   & 0             \\ \hline
\end{tabular}
\caption{Example of DouBan DuShu dataset \label{douban}}
\end{table}

\subsection{Chiphell}
Chiphell \footnote{available on: https://github.com/JaniceZhao/Chinese-Forum-Corpus.git} is a web portal where netizens share their views to news and discuss within groupuscule. We have collected discussion forums from several subjects, such as computer hardware, motors and clothes. There are more than 50 thousand discussions in the corpus. Similar to the DouBan DuShu corpus, most of the sentences collected from Chiphell are informal Chinese and some of them are in particular domains. An example from each subject is shown in table \ref{chh}.

\begin{table}[h!]
\centering
\begin{tabular}{l|p{6cm}|p{6cm}}
\hline
Subject            & Topic                          & Example                      \\ \hline
News               & \begin{CJK*}{UTF8}{gbsn}美机场航空业希望修改客机降落的Emoji表情：机头朝下不吉利\end{CJK*} & \begin{CJK*}{UTF8}{gbsn}那我还说改完的意思是无限复飞呢,飞到没油不又gg了\end{CJK*}    \\
Computer Hardware & \begin{CJK*}{UTF8}{gbsn}请问现在大船货除开3610还有其他性价比的大船大容量吗\end{CJK*}    & \begin{CJK*}{UTF8}{gbsn}我1T的PM1633。。卖1300都木有人接\end{CJK*}       \\
Mobile Phones      & \begin{CJK*}{UTF8}{gbsn}努比亚X 综合讨论帖\end{CJK*}                     & \begin{CJK*}{UTF8}{gbsn}MIX3辣鸡被友商各种吊打\end{CJK*}                \\
Clothes            & \begin{CJK*}{UTF8}{gbsn}程序媛的皮艺生活\end{CJK*}                       & \begin{CJK*}{UTF8}{gbsn}花点时间在复杂又感兴趣的事情上是一件快乐又有成就感的体验\end{CJK*} \\ \hline
\end{tabular}
\caption{Example of Chiphell dataset \label{chh}}
\end{table}

\subsection{Data Pre-processing}

In addition to the raw dataset, we extracted the comments and preprocessed them to provide a clean, formal formatted and comprehensive Chinese corpus. After carefully investigate the raw text, mainly three preprocessing methods are applied:
\begin{enumerate}
\item convert Traditional Chinese to Simplified Chinese 
\item remove over-short comments (less than 4 characters)
\item add identifier to special characters, such as special signs, English words and emoticons
\end{enumerate}

\section{Experiments}

To further explore the informal Chinese corpus, we calculate the proportion of informal words in the corpus. The experiment is conducted on Weibo News \cite{hu2015lcsts}, Sougou News, People's Daily \cite{yu2001guideline} and the LSICC. We manually collected 70 informal words as the benchmark, which covers both spoken Chinese words and Chinese network slang words.

We counted the frequencies of informal words and the number of total words to calculate the proportion of the informal words in the whole corpus. As shown in table \ref{proportion}, the LSICC has the highest proportion of the informal words, which is more than two times the second highest one, the Weibi News. Noted that the more formal the media is, the lower the proportion of the informal words in it.

\begin{table}[h!]
\centering
\begin{tabular}{l|l|l|l}
\hline
Corpus         & Informal Words & Total Words & Proportion \\ \hline
LSICC          & 621807         & 705231306   & 8.82\textperthousand     \\
Weibo News     & 46831          & 125082112   & 3.74\textperthousand     \\
Sougou News    & 1238           & 14160148    & 0.87\textperthousand     \\
People's Daily & 25             & 3482887     & 0.07\textperthousand     \\ \hline
\end{tabular}
\caption{Proportion of the informal words in each corpus \label{proportion}}
\end{table}

The result indicated that the gap between the language that the real-world natural language models deal with the existing corpora is significant. Using the vector representations extracted from the corpus of formal Chinese as the word embedding may attribute to poor performance.

\section{Conclusions and Future Work}

We constructed a large-scale Informal Chinese dataset and conducted a basic words frequency statistic experiment on it. Compared to the existing Chinese corpus, LSICC is more typical dataset for real-world natural language processing tasks, especially for sentiment analysis. As a next step, we should conduct embedding extraction Chinese words segmentation and sentiment analysis on LSICC. Meanwhile, as the raw information, such as the usernames and book names is kept, LSICC can also be used to build recommendation systems and explore social network.

\bibliographystyle{unsrt}
\bibliography{references}

\end{document}